# EMAIL SPAM DETECTION USING HIERARCHICAL ATTENTION HYBRID DEEP LEARNING METHOD

Sultan ZAVRAK[1*] and Seyhmus YILMAZ[1]

Sultan ZAVRAK

[1]Department of Computer Engineering, Duzce University, Duzce/TURKEY

sultanzavrak@duzce.edu.tr

**ORCID ID:** 0000-0001-6950-8927

Seyhmus YILMAZ

[1]Department of Computer Engineering, Duzce University, Duzce/TURKEY

seyhmusyilmaz@duzce.edu.tr

***ORCID ID:** 0000-0001-9987-2797*

***Corresponding Author:**

Sultan ZAVRAK

Department of Engineering,
Faculty of Engineering, Duzce University, 81620, Duzce / TURKEY
Phone: +90 (380) 542 1036
E-mail:  sultanzavrak@duzce.edu.tr

# EMAIL SPAM DETECTION USING HIERARCHICAL ATTENTION HYBRID DEEP LEARNING METHOD

**Abstract**

Email is one of the most widely used ways to communicate, with millions of people and businesses relying on it to communicate and share knowledge and information on a daily basis. Nevertheless, the rise in email users has occurred a dramatic increase in spam emails in recent years. Processing and managing emails properly for individuals and companies are getting increasingly difficult. This article proposes a novel technique for email spam detection that is based on a combination of convolutional neural networks, gated recurrent units, and attention mechanisms. During system training, the network is selectively focused on necessary parts of the email text. The usage of convolution layers to extract more meaningful, abstract, and generalizable features by hierarchical representation is the major contribution of this study. Additionally, this contribution incorporates cross-dataset evaluation, which enables the generation of more independent performance results from the model's training dataset. According to cross-dataset evaluation results, the proposed technique advances the results of the present attention-based techniques by utilizing temporal convolutions, which give us more flexible receptive field sizes are utilized. The suggested technique's findings are compared to those of state-of-the-art models and show that our approach outperforms them.

**Keywords:** Hierarchical Attentional Hybrid Neural Networks, email spam detection, Natural Language Processing, FastText, attention mechanisms

1. Introduction

In the age of information technology, information sharing has become more straightforward and rapid than ever. In many parts of the world, numerous platforms are accessible for individuals to share information anyplace [1]. Email is unquestionably the most prevalent and cost-efficient method of exchanging information with people who own electronic equipment. An email has found widespread use in a wide variety of human activities, including but not limited to communicating significant posts within an organization, inter-organization, or worldwide; job recruitment processing, advertisements, health care communications, bank transactional information, and inter/intra-organizational correspondence [2]. Due to the enormous benefits of email, its use has been harmed by the prevalence of fake emails and occasionally deceitful emails that must be immediately identified and isolated via what is generally known as a spam detection system. A spam filtering process is required, which entails using software methods to distinguish between spam and non-spam email, allowing for the prevention of spam mail from reaching users' inboxes [3]. While numerous methods have been implemented to eliminate the threat of spam or significantly reduce the amount of spam directed at internet users worldwide, the reported accuracy indicates that additional work in this area is necessary [4].

As in various fields, the deep learning techniques are proven to have superior performance to other machine learning algorithms in this area [5]. In addition to this, deep neural network techniques play an important part in Natural Language Processing (NLP) with the representation of word vectors learned. It is straightforward to represent words in terms of continuous, fixed-length, and dense feature vectors, thereby representing semantic word relationships: related words are spatially close to one another in the vector space [6].

In many natural language applications that include email spam classification, the contextual significance of sentences and words is not considered by the suggested architectures. Also, the information on the text structure in the model is not efficiently included. Informative or qualitative words and sentences are not used in many techniques; nevertheless, several words may be more informative than others in a text. Additionally, RNNs are generally the only deep learning method used in such techniques.

Recently, attention mechanisms are a new tendency in natural language processing to build information dependencies regardless of their distance between words in the input sequences. For instance, the authors of [7] propose a hierarchical deep learning model for document categorization that makes use of attention mechanisms to approximate the document's hierarchical structure. The fundamental premise of this approach is that every section of a document is not equally related to representing it [8]. In addition to that, identifying relevant parts includes modeling not only the occurrence of words in the text but also the significance and interactions between words.

In this study, a novel technique based on CNN and gated recurrent unit (GRU) [9] with an attention mechanism for email spam classification has been developed. More specifically, we propose a hierarchical model for email spam detection based upon CNN, GRU [9] attention mechanisms, and hidden units to advance the architecture results by selectively concentrating the model on necessary sections of the text emails when training the model. A hierarchical model called Hierarchical Attentional Hybrid Neural Networks (HAN) is implemented to better represent email text structure as described in [8]. In this model, temporal convolutions [10] are also used in order to provide more flexibility in receptive field sizes. In addition to this, the proposed approach is tested by making a comparison of its performance with recent advanced methods, and the results are evaluated using various metrics show better performance.

The following is the organization of the article. Section two summarizes published research on spam classification. The third section discusses the theoretical foundations of the proposed approach. The fourth section contains information about the experimental methodology, the description of the datasets, and the proposed method's cross-dataset evaluation results, as well as a comparison to previous studies. Concluding remarks are stated in the final section.

2. Related work

Email classification is a promising and widely used technique for spam email detection (e.g., in mobile social networks [11]). Numerous machine learning techniques, including semi-supervised and supervised learning algorithms, have been investigated to discriminate between suspicious and legitimate emails.

2.1. Supervised learning algorithms

Naive Bayes (NB), k-nearest neighbor (KNN), Support Vector Machine (SVM), ensemble learning, decision tree, and many other supervised machine learning algorithms have been studied in the literature. For example, in the context of email classification for spam management, the author of [12] proposes a hardware system based on NB classifier. The authors demonstrated a word-serial NB classifier that uses the Logarithmic Number System (LNS) to reduce computational complexity and non-iterative binary LNS recoding using a look-up table technique. According to the findings, their method can handle a large number of emails in a short period of time.

The authors of [13] introduced a spam-behavioral detection system and implemented a Fuzzy Decision Tree based spam filter system, which can calculate Information Gain to examine and pick behavior features of emails. To recognize spam emails, a new classification approach based on a decision tree is implemented and ensemble learning is illustrated in [14]. In the majority of cases, the assessments on a public dataset illustrated that the suggested approach exceeded benchmark methods such as SVM, C4.5, KNN, and Naive Bayes.

The authors of [15] illustrated a method for generating spam filters using KNN and SVM. They especially proposed an offline implementation that employs a pre-classified email dataset and the k-Nearest Neighbor (kNN) approach for the learning stage. This model can execute a continuous update to the dataset and the most frequent word list that is present in the documents throughout the evaluation.

The authors of [16] suggest the make use of SVM to classify email messages as valid or spam by making a comparison of SVM with three classification methods: Boosting decision trees, Rocchio and Ripper. These four methods are evaluated on two different datasets, where Support Vector Machines are performed the best when managing binary features.

The authors in [17] demonstrated for the first time that online SVMs did achieve state-of-the-art classification performance on large benchmark datasets for online spam filtration. They showed that a Relaxed Online SVM could achieve roughly the same results at a much lower computational cost. Experiments on blog spam detection, email spam detection, and splog detection were used to validate their findings.

Zhan et al. [18] proposed a stochastic learning method that uses weak estimators to model unusual emails in a dynamic context. A multivariate Bernoulli NB classifier was employed during the training phase. The results of the experiments indicate that detecting suspicious emails is both possible and effective.

In [19], the authors investigated the use of an inductive learning technique based on the Group Method of Data Handling to detect spam communications by automatically detecting content features that could successfully distinguish spam from legitimate emails. In comparison to other algorithms such as neural networks and Naive Bayes, their solution has a higher accuracy of spam detection and requires less training time, with false-positive rates as low as 4.3 percent.

In a large-scale empirical study, Ouyang et al. [20] investigated the effectiveness of using packet and flow data based on decision trees and Rulefit to detect spam emails in an enterprise. Numerous further studies on this subject can be found in [21]–[28] and [29].

**2.2. Semi-Supervised learning algorithms**

Due to the large amount of labeled data required for supervised learning, semi-supervised learning was developed to classify both unlabeled and labeled data. For example, the authors of [30] demonstrated how to improve classification accuracy by combining an SVM and a semi-supervised classifier. To classify a user's emails, the SVM is trained on labeled public domain emails, whereas the semi-supervised classifier uses these emails as the training set and propagates the label information to unlabeled emails via feature space distribution.

Then the authors in [31] suggested a semi-supervised classifier ensemble with the goal of labeling users' emails and facilitating the tuning procedure. This semi-supervised ensemble was tested to assist Support Vector Machine in classifying users' emails with high accuracy.

Gao et al. [32] introduced a semi-supervised solution for detecting picture spam emails named the regularized discriminant EM algorithm. In comparison to fully supervised learning methods, they discovered that the expense of collecting sufficient labeled data for training was too high for fully supervised learning. In contrast, their method could recognize invalid emails while also training a classification model using a large amount of unlabeled data and a small amount of labeled data.

In [33], a particular scenario for semi-supervised spam filtration used by the authors: that is when a great deal of training data is obtainable, but only a few true labels can be acquired for that data. As a result of this, the authors illustrated two spam filtering methods for this situation, both beginning with a cluster of training emails. Their technique may give superior results than those beforehand published state-of-the-art semi-supervised techniques on small-sample spam filtration in the evaluation.

By combining disagreement-based semi-supervised learning and multi-view data, the authors of [34] propose an effective classification architecture. The authors then proceeded to extend their previous study and look at its impact in an IoT environment [35]. They collaborate with a real-world IT company to test the technique in a real-world network environment. Furthermore, they highlight several limitations as well as open challenges in this field.

Other related works in the area of semi-supervised learning in email classification can be found [36]–[39], as well as several surveys on spam filtration [40]–[42].

### 2.3. Deep-based learning methods

Recently, the most advanced results for spam email detection have been obtained by deep learning-based architectures. In [43], the authors use the present machine learning algorithms such as CNN, LSTM, Naive Bayes, and SVM to recognize and classify e-mail messages. The LSTM architecture gives better performance than other machine learning methods with a maximum score of 98.4% accuracy according to experimental results. As well as this, CNN, NB, and SVM models achieve an accuracy of 96.20%, 98.00%, and 97.50% respectively. Furthermore, 92.50% recall, 97% precision, and an F1 score of 95% are obtained by Support Vector Machine architecture and F1 score of 96%, 96.50% precision, and 95% recall is obtained by NB architecture.

The authors of [44] generated a 3-fold classifier using an artificial neural networks model (ANNs). The highest accuracy of 99.57% and the highest F1 score, recall, and precision of 99.68% is obtained by their architecture when the Spambase dataset is used. In order to discriminate spam emails from ham, a new technique that includes transformers and deep learning methods using the self-attention mechanism has been used by researchers in NLP downstream applications in recent years. When pre-training, transformers consist of a restricted selection of models [45]. BERT gives better results by employing a masked language architecture to utilize pre-trained deep bidirectional representations (Devlin et al. 2018).

In order to distinguish spam emails from ham, the authors of [45] developed a BERT-based architecture. The authors implemented and trained a spam recognition architecture by using two different datasets, which relied on a pre-trained BERT. Some machine learning models such as KNN, SVM, logistic regression, and RF are used to compare the results of the models. In addition to this, a number of evaluation metrics, for example, F1 score, recall, and precision are used to analyze the performance of the models. Two publicly available datasets are used. When the EN dataset is used, the logistic regression technique reaches the highest F1 score of 97.84%, recall score of 97.83%, and the precision score of 97.86%. On the other hand, the detection results are obtained with "the spam or not spam dataset": 95.95% precision, 96% recall, and an F1 score of 95.92%.

A BERT-based spam recognition technique is proposed by the [47]. To implement the proposed method, the Spam Filter Dataset from Kaggle and Spambase from the UCI Machine Learning Repository is used. They also compare the proposed BERT-based network with bidirectional LSTM (BiLSTM) architecture, NB-based architecture, and a KNN-based architecture. According to experimental results, the BERT-based architecture gives better performance than other machine learning methods with the highest score of 98.67% accuracy and 98.66% F1. Bidirectional long short-term memory architecture, the NB architecture, and the KNN architecture reach an accuracy of 96.43%, 94.69%, and 92.92% respectively.

A pre-trained BERT uncased architecture is developed in [48]. The developed model consists of three fully connected linear layers with batch normalization layers, the log softmax, and four dropout layers. The proposed architecture reaches an accuracy of 98% when SpamText, LS, and the SA datasets are used and an accuracy of 97% when the EN dataset is used. An F1 score of 90% is obtained in the architecture while each architecture utilizes its relevant datasets. The universal model is implemented by using four datasets and employed hyperparameters from each architecture. The model archives an overall accuracy of 97%, with an F1 score of 96% on the combined dataset.

3. Hierarchical Attentional Hybrid Neural Networks

The proposed approach is the combination of convolution layers, attention mechanisms, and Gated Recurrent Units [8]. The proposed method is shown in Figure 1. The proposed method's first layer is a pre-processed word embedding layer, represented by black circles in Figure 1. The next layer is a CNN layer stack that includes convolution layers with multiple filters (variable window sizes) and feature maps. A number of trials with temporal convolution layers with dilated convolutions have also been carried out, yielding promising results. Dropout is also used to achieve regularization. In the third layer, an attention mechanism is applied to the word-level context vector by a word encoder. Sentence encoders apply attention to a sentence-level context vector in a sequential manner. To produce the output probability distribution over the classes, the final layer contains a SoftMax activation function.

The hierarchical representation of CNN is used to detect more generalizable, meaningful, and abstract features. By combining convolution layers in various filter sizes with both sentence and word encoders in a hierarchical architecture, the proposed method detects more rich features and improves generalization performance in email spam classification. FastText (FT) is used in the word embedding initialization by taking into consideration sub-word information to gain representations of more uncommon words [49].

In this paper, two variants of the proposed frameworks have been investigated. As shown in Figure 1, There is a basic version and a TCN [10] layer implementation version. The goal is to use a combination of regular and dilated convolutions with residual connections to simulate RNNs with very large memory sizes. In long sequences, dilated convolutions are preferred because they allow for an exponentially larger receptive field in convolution layers.

In more technical terms, for the input of a 1-D sequence $x \in \mathbb{R}^n$ and a filter $f: \{0, \dots, k-1\} \to \mathbb{R}$, on element s of the sequence, the dilated convolution function $F$ is defined as

$$F(s) = (x *_d f)(s) = \sum_{i=0}^{k-1} f(i) \cdot x_{s - d \cdot i} \qquad 1$$

where $d$ is the dilatation factor, $k$ denotes the filter size, and $s - d \cdot i$ denotes the direction of previous information. Therefore, dilation is equal to introducing a constant step among each pair of neighboring filter maps. When $d$ equals 1, a dilated convolution becomes a regular one. The make use of larger dilation allows an output at the top level to represent a wider range of inputs, increasing the receptive field.

The proposed architecture accounts for the fact that the various sections of a text do not contain any information that is related to each other. Modeling the interactions between the words, not just their isolated appearance in the document, is also part of defining the related parts. As a result, there are two levels of attention mechanisms in the network [50]. The first structure is at the sentence level, while the second is at the word level, allowing the network to give more or less attention to particular sentences and words when generating the text representation.

The approach is divided into two components: a word sequence encoder and a layer of attention at the word level; and a sentence encoder and a layer of attention at the sentence level. The network employs bidirectional GRU [50] to generate word annotations by summarizing input in both ways in the word encoder. As a result, the annotation integrates contextual information. When creating the text's representation, the model can pay more or less attention to specific words and sentences depending on the attention levels. [7].

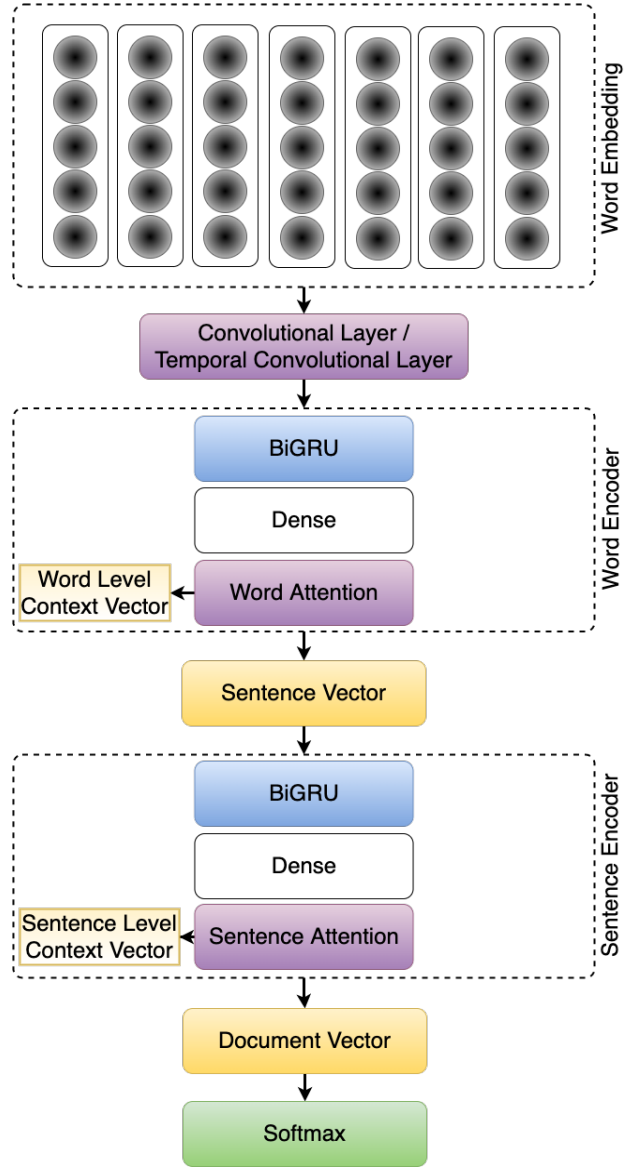

**Figure 1.** After the embedding layer, the proposed method for email classification includes a CNN layer. A variant that contains a temporal convolutional layer [8] after the embedding layer.

Given a sentence that has the words $w_{it}, t \in [0, T]$, Using an embedding matrix $W_e$, a bidirectional GRU has a forward GRU $\overrightarrow{f}$ that reads the sentence $s_i$ from $w_{i1}$ to $w_{iT}$, as well as a reverse GRU $\overleftarrow{f}$ that reads the sentence $s_i$ from $w_{iT}$ to $w_{i1}$:

$$x_{it} = W_e w_{it}, t \in [1, T] \qquad 2$$

$$\overrightarrow{h_{it}} = \overrightarrow{GRU}(x_{it}), t \in [1, T] \qquad 3$$

$$\overleftarrow{h_{it}} = \overleftarrow{GRU}(x_{it}), t \in [T, 1] \qquad 4$$

When a word $w_{it}$ is given, the annotation is acquired by concatenating the forward hidden state and the backward hidden state, i.e., $h_{it} = [\overrightarrow{h_{it}}; \overleftarrow{h_{it}}]$, which contains a summary of the information contained in the entire sentence. When a sentence contains words that are critical to understanding the meaning of the sentence, the attention mechanism is activated, and the representation of those informative words is aggregated into a sentence vector. In particular,

$$u_{it} = \tanh(W_w h_{it} + b_w) \qquad 5$$

$$\alpha_{it} = \frac{\exp(u_{it}^\top u_w)}{\sum_t \exp(u_{it}^\top u_w)} \qquad 6$$

$$s_i = \sum \alpha_{it} h_{it} \qquad 7$$

The architecture determines a word's significance by comparing $u_{it}$ to a word-level context vector $u_w$ and then obtains a normalized significance weight $\alpha_{it}$ through a softmax function. Following that, based on the weights, the model calculates the sentence vector $s_i$ as a weighted sum of the word annotations. Throughout the process of training, the word context vector $u_w$ is initialized at random and cooperatively trained.

The attention of the sentence is calculated using the sentence vectors $s_i$, and the document vector in the following way:

$$\overrightarrow{h_{it}} = \overrightarrow{GRU}(s_i), i \in [1, L] \qquad 8$$

$$\overleftarrow{h_{it}} = \overleftarrow{GRU}(s_i), i \in [L, 1] \qquad 9$$

The suggested method concatenates $h_{it} = [\overrightarrow{h_{it}}; \overleftarrow{h_{it}}]$ $h_i$ which summarizes the neighbor sentences surrounding sentence $i$ while staying focused on sentence $i$. Using an attention mechanism, the method begins by creating a sentence-level context vector $u_s$, and using it to determine the significance of the sentences to reward sentences that are relevant to accurately classifying a document,

$$u_{it} = \tanh(W_s h_i + b_s) \quad \text{10}$$

$$\alpha_{it} = \frac{\exp(u_i^T u_s)}{\sum_i \exp(u_i^T u_s)} \quad \text{11}$$

$$v = \sum \alpha_i h_i \quad \text{12}$$

$v$ denotes the document vector, which summarizes all of the information in a document in the previous equation. Thus, at the sentence level, the context vector $u_s$ can be randomly initialized and jointly learned during the training process. The output of the sentence attention layer is fed into a fully connected SoftMax layer. It generates a probability distribution for each of the classes.

## 4. Experiments and Results

### 4.1. Datasets Description

This study experimented on five widely used datasets: TREC 2007 (TR) [51], GenSpam (GS) [52], SpamAssassin (SA) [53], Enron (EN) [54], and Ling Spam (LS) [55]. These datasets are available to the public for free.

Table 1. Databases content breakdown

| Dataset | Ham | Spam | Total |
| --- | --- | --- | --- |
| TR | 25217 (%33.49) | 50071 (%66.51) | 75288 |
| GS | 9186 (%23) | 30761 (%77) | 39947 |
| SA | 4144 (%68.65) | 1892 (%31.35) | 6036 |
| EN | 16544 (%49.16) | 17110 (%50.84) | 33654 |
| LS | 2412 (%83.37) | 481 (%16.63) | 2893 |

To correctly compare results with previous works, the dataset preprocessing methodology described in [51] is adopted. When it was first released, the SA dataset was broken down into five separate files: easy-ham, easy-ham-2, hard-ham, spam, and spam-2. It is worth noting that the first two files contain emails that are easily distinguishable from spam emails and do not contain spammy signatures. There are ham emails in the third file that are more in line with standard spam. In the end, the fourth and fifth file includes emails that were obtained from legitimate sources and did not fall into the spam traps. All of the emails were combined into two distinct files, one for ham and one for spam. Initially, the GS dataset was broken down into five separate folders: train_GEN, train_SPAM, test _GEN, test_SPAM, adapt_GEN, and adapt_SPAM, which contained ham (GEN) and spam for every stage of training, testing, and validation. When tests were run on the same dataset, the original data split was used; however, when cross-dataset training was performed, all emails from the folders were united into two distinct files designated spam and ham and then tested on those files.

In the beginning, the Enron dataset was divided into six folders, each of which contained spam and ham email messages. All emails were categorized into two folders: ham and spam. Even though the LS dataset initially contained multiple versions of the data, the bare version, which was unprocessed, was used for this study. It is divided into ten sections, each of which contains some form of spam or ham. It was determined that the original split of data would be used when multiple experiments were conducted on the same dataset; however, when experiments were conducted on different datasets, the emails from the folders were merged into two files, which were then labeled spam and ham, respectively. Lastly, the complete TR dataset was used, in which each email is labeled as either spam or ham.

The distribution of emails by class is shown in Table 1, with the first class receiving the most emails. Spam email is prevalent in the majority of the sample (owing to its ease of collection), excluding in SA, where the ratios are inverted, and EN, where it is a nearly balanced dataset.

Table 2. Vocabulary sizes and statistics for superficial features

| Dataset | Vocabulary | | | | Average per email | | |
|---|---|---|---|---|---|---|---|
| | Words | Links | Emoticons | Voc Words/#Emails | Words | Links | Emoticons |
| TR | 5231250 | 11193 | 102 | 69.48 | 264.53 | 0.51 | 8 |
| GS | 93427 | 0 | 113 | 2.34 | 52.03 | 0 | 0.13 |
| SA | 142461 | 6107 | 162 | 23.6 | 172.35 | 2.33 | 5.79 |
| EN | 158652 | 5739 | 158 | 4.71 | 117.59 | 0.33 | 1.19 |
| LS | 58950 | 1234 | 126 | 20.38 | 239.35 | 0.64 | 2.79 |

Table 2 shows the statistics for the three superficial features tested: links, words, and emojis/emoticons. The size of the vocabulary associated with each feature in the dataset (the number of unique links, words, and emojis/emoticons) is indicated in the second to fourth columns. The fifth column indicates the percentage of emails that contain unique words (the number of emails divided by the size of the word vocabulary). The higher this number is, the more diverse the text in the dataset is. The averages for each feature per email in the dataset are found in columns six to eight. As seen in this table, TR has the greatest word and link vocabularies, implying that its content is more diverse than those of the other datasets.

The fact that its emails are on average the largest and have the most emojis of any dataset backs this up. Despite its size, the average length of emails in the GS dataset is the smallest., resulting in a moderate word vocabulary. This dataset's emails have no links and have the fewest emoticons. Despite its small size, SA's word, emoticon, and link vocabularies are larger. Emails in this dataset, on average, have the most links and the second most emoticons. In terms of size, the EN is the third biggest dataset, with a medium vocabulary of words, a number of links, and, a large number of emoticons per email message. Finally, while the LS dataset is the smallest, it contains the second-most emails, resulting in a moderate link and word vocabulary. Additionally, on average, its emails contain a number of links and emoticons. Generally, emails in the datasets had few links and emojis, as well as sparse vocabularies for these features, notably emojis, which should be taken into account given the restricted amount of symbols accessible for this feature. The number of words, on the other hand, varies significantly, as do the sizes of the word vocabulary.

**4.2. Evaluation Metrics**

The binary classification results fall into four categories: 1) True Positive (TP): Spam was correctly classified. 3) False Positive (FP): Misclassification of a normal instance 4) True Negative (TN): Correct classification of normal instances. Aside from that, previous metrics can be used to derive subsequent metrics. [56].

Classification accuracy is a basic evaluation metric. The model's accuracy is calculated as follows:

$$Accuracy = (TP + TN)/(TP + FP + TN + FN) \qquad 13$$

The precision measures how many of the positive predictions are true positives.

$$Precision = TP/(TP + FP) \qquad 14$$

The recall is the proportion of correctly predicted positive instances among all positive instances.

$$Recall = TP/(TP + FN) \qquad 15$$

F-measure is the harmonic mean of precision and recall; the traditional F-measure is computed as:

$$F = 2\frac{Recall \times Precision}{Recall + Precision} \qquad 16$$

which is also known as the F1 measure since here the weight of precision and recall is equal.

Receiver Operating Characteristics (ROC): The area under the receiver operating characteristic curve (AUC) metric is widely used as a de facto criterion for classifier effectiveness evaluation [57], [58]. AUC is a metric for determining how much a positive class instance ranks higher than a negative class instance at random after being sorted by classification probabilities. Although AUC has a range of possible values between 0 and 1, a random classifier would generate the ROC curve's diagonal, resulting in an AUC of 0.5, which is the baseline performance [59].

### 4.3. Experimental Results and Discussion

The proposed model is evaluated using five email classification datasets in this study. To build and test the model, two types of experiments were undertaken: one in which the same dataset was employed for both training and testing, and the other in which one dataset was employed to build the model, and the remaining ones were used as test datasets. Cross-validation with the TR, SA, EN, and LS datasets is performed tenfold using the same dataset configuration. Cross-validation is stratified at random for the first three databases, however, for LS, the original data split is utilized.

GS makes use of the original data segmentation into training, test, plus validation dataset. Throughout the tenfold cross-validation, an independent vocabulary consisting of nine folds was extracted from the training section. Following that, both the training and remainder of the test fold are vectorized using this vocabulary. Vectorization of the word embeddings with the vocabulary from the training section was performed using the FT model.

As previously stated, the cross-dataset setup initially combined all emails from each dataset into two files, including spam and ham. Following that, vocabulary was extracted from each training dataset. Finally, using word embeddings with FT, this vocabulary was used to perform transformations on all test datasets.

The email spam detection task, by definition, involves uneven data distributions, in which one class is more prevalent than the other (as displayed in Table 1). In this scenario, accuracy measures are not suggested for evaluating a model's classification performance, as they are likely to be biased toward the dominant class. In our situation, we evaluated the performance of the model using the AUC which is a widely used metric for email categorization [9]. We also give the evaluation results in terms of accuracy, recall, precision, and F1 metrics.

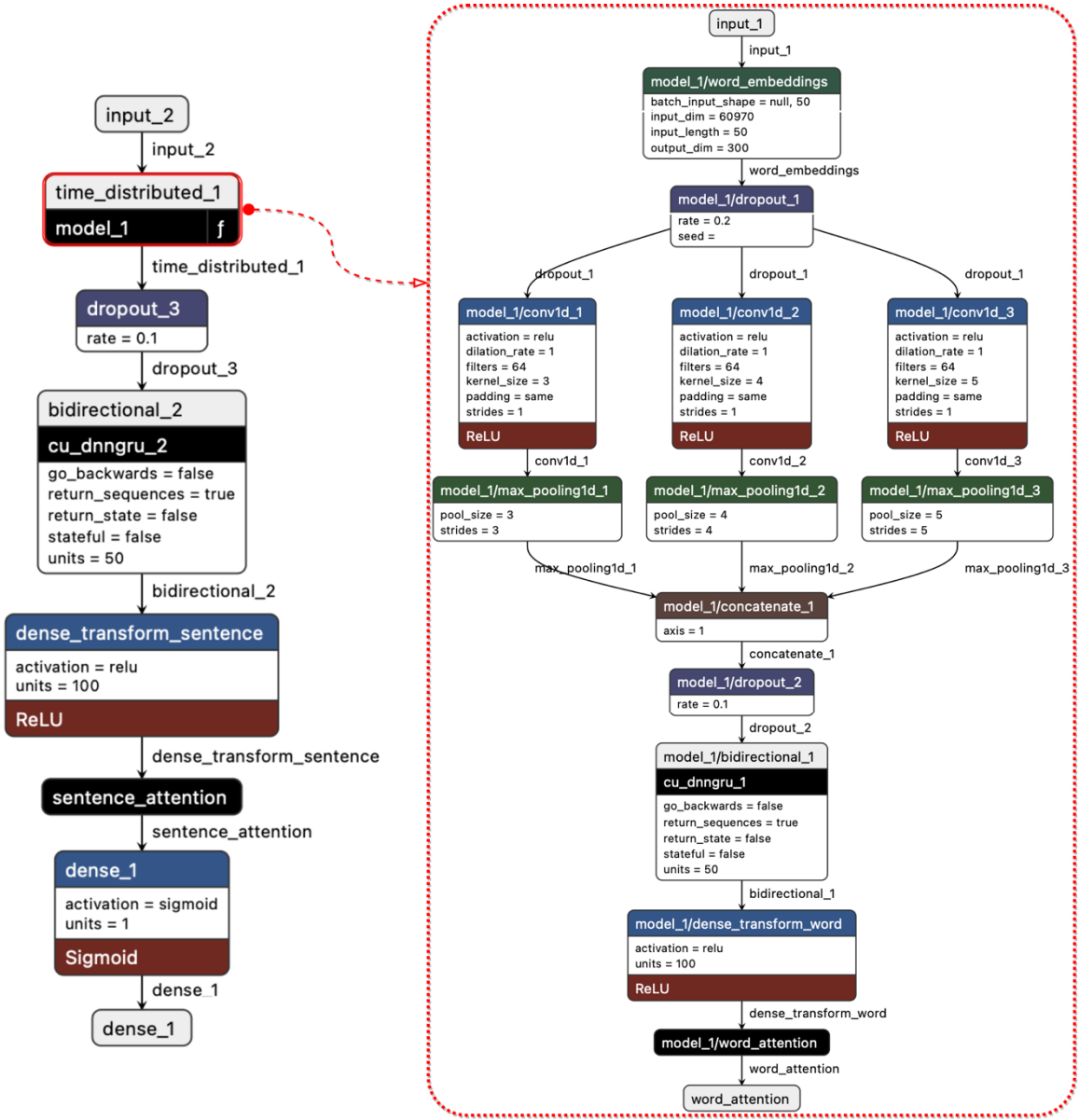

Figure 2. The visualization of the proposed network model architecture

Figure 2 depicts a visual representation of the proposed network model architecture. The internal structure of the *model_1* component shown on the left of the figure is depicted on the right. The figure depicts information such as the type of the layer, the layer's name, the layer's activation function, the number of neurons in each layer, the number of filters, the kernel size, the strides, padding, data format, and dilation rate of the convolutional layers, etc. In other words, it includes almost all the details of the model for reproducibility of this study.

Table 3. All results of the FT+HAN method

| Train | | Test | Accuracy | Precision | Recall | F1 | AUC |
|---|---|---|---|---|---|---|---|
| TR | | TR | 0.992 | 0.989 | 0.999 | 0.994 | 0.999 |
| | | GS | **0.784** | **0.786** | 0.989 | **0.876** | 0.675 |
| | | SA | 0.743 | 0.551 | 0.972 | 0.703 | **0.879** |
| | | EN | 0.621 | 0.573 | **0.990** | 0.726 | 0.864 |
| | | LS | 0.410 | 0.217 | 0.981 | 0.356 | 0.832 |
| GS | | TR | **0.761** | **0.745** | 0.975 | **0.845** | 0.779 |
| | | GS | 0.954 | 0.953 | 0.957 | 0.955 | 0.991 |
| | | SA | 0.529 | 0.399 | **0.994** | 0.569 | 0.817 |
| | | EN | 0.632 | 0.582 | 0.982 | 0.731 | 0.804 |
| | | LS | 0.714 | 0.365 | 0.975 | 0.531 | **0.942** |
| SA | | TR | **0.862** | **0.845** | 0.970 | **0.903** | 0.899 |
| | | GS | 0.830 | 0.829 | **0.981** | 0.899 | 0.718 |
| | | SA | 0.955 | 0.893 | 0.978 | 0.933 | 0.987 |
| | | EN | 0.617 | 0.573 | 0.968 | 0.720 | 0.789 |
| | | LS | 0.313 | 0.192 | 0.979 | 0.322 | 0.750 |
| EN | | TR | 0.800 | **0.936** | 0.750 | 0.833 | 0.892 |
| | | GS | 0.795 | 0.909 | 0.817 | **0.860** | 0.836 |
| | | SA | 0.842 | 0.712 | 0.834 | 0.768 | 0.903 |
| | | EN | 0.958 | 0.981 | 0.937 | 0.958 | 0.989 |
| | | LS | **0.916** | 0.703 | **0.857** | 0.772 | **0.957** |
| LS | | TR | **0.723** | **0.798** | 0.781 | **0.789** | 0.805 |
| | | GS | 0.614 | 0.761 | 0.727 | 0.744 | 0.480 |
| | | SA | 0.574 | 0.409 | 0.815 | 0.545 | 0.734 |
| | | EN | 0.718 | 0.662 | **0.912** | 0.767 | **0.830** |
| | | LS | 0.980 | 0.933 | 0.948 | 0.940 | 0.997 |

We split documents into sentences and tokenize them individually. The dimension of the word embeddings is 200, the batch size is 64, and the Adam optimizer with a learning rate of 0.001 is used.

Python libraries NLTK [60], NumPy [61], sci-kit-learn [62] and Keras with Tensorflow backend were used to develop the processing and classification algorithms. Experiments were conducted on a Linux workstation equipped with a 2.6 GHz Intel Core i7 9750H processor and 32 GB of RAM.

Table 4. AUC Results Comparison

|  |  |  | TRAIN |  |  |  |  | SD AVRG | CD AVRG |
|---|---|---|---|---|---|---|---|---|---|
|  |  |  | TR | GS | SA | EN | LS |  |  |
| TEST | W2V+SVM | TR | *0.95* | 0.75 | 0.76 | 0.71 | 0.71 | 0.954 (0.023) | 0.7505 (0.085) |
|  |  | GS | 0.58 | *0.94* | 0.8 | 0.75 | 0.74 |  |  |
|  |  | SA | 0.71 | 0.84 | *0.93* | 0.65 | 0.71 |  |  |
|  |  | EN | 0.7 | 0.7 | 0.67 | *0.96* | **0.77** |  |  |
|  |  | LS | **0.85** | **0.95** | **0.89** | **0.77** | *0.99* |  |  |
|  | W2V+LR | TR | *0.95* | 0.75 | 0.76 | 0.71 | 0.71 | 0.956 (0.021) | 0.75 (0.085) |
|  |  | GS | 0.58 | *0.95* | 0.8 | 0.75 | 0.71 |  |  |
|  |  | SA | 0.73 | 0.83 | *0.93* | 0.65 | 0.71 |  |  |
|  |  | EN | 0.69 | 0.71 | 0.67 | *0.96* | **0.76** |  |  |
|  |  | LS | **0.85** | **0.95** | **0.89** | **0.79** | *0.99* |  |  |
|  | W2V+KNN | TR | *0.96* | 0.72 | 0.72 | 0.73 | 0.68 | 0.952 (0.025) | 0.7155 (0.085) |
|  |  | GS | 0.55 | *0.93* | 0.75 | **0.79** | 0.69 |  |  |
|  |  | SA | 0.64 | 0.83 | *0.92* | 0.72 | 0.66 |  |  |
|  |  | EN | 0.67 | 0.69 | 0.61 | *0.97* | 0.67 |  |  |
|  |  | LS | **0.72** | **0.93** | **0.85** | 0.69 | *0.98* |  |  |
|  | W2V+RF | TR | *0.95* | 0.66 | 0.63 | 0.67 | 0.65 | 0.93 (0.033) | 0.693 (0.076) |
|  |  | GS | 0.56 | *0.88* | 0.72 | 0.69 | 0.67 |  |  |
|  |  | SA | 0.69 | 0.74 | *0.91* | 0.61 | **0.69** |  |  |
|  |  | EN | 0.65 | 0.67 | 0.63 | *0.95* | 0.68 |  |  |
|  |  | LS | **0.77** | **0.9** | **0.81** | **0.77** | *0.96* |  |  |
|  | FT+HAN (This study) | TR | 0.999 | 0.779 | **0.899** | 0.903 | 0.805 | 0.9926 (0.005) | 0.80645 (0.105) |
|  |  | GS | 0.675 | 0.991 | 0.718 | 0.836 | 0.480 |  |  |
|  |  | SA | **0.879** | 0.817 | 0.987 | 0.836 | 0.734 |  |  |
|  |  | EN | 0.864 | 0.804 | 0.789 | 0.989 | **0.83** |  |  |
|  |  | LS | 0.832 | **0.942** | 0.75 | **0.957** | 0.997 |  |  |

The performance results in terms of accuracy, precision, recall, and F1 are shown in Table 3. In addition to this, Table 4 illustrates the experiment result of our method utilizing FT and the described features, links, words, emojis /emoticons, and Word2Vec that were used by the authors of [63]. The tables are divided into 5x5 blocks, with the AUC values for each classification network and feature in each block. The datasets used for training are represented by the columns in the block, and the datasets used for testing are represented by the rows. In a block containing results from the same dataset, the AUC values are italicized in the diagonal. The best values in a row are bolded. The standard deviation and averages of all the results for a single training dataset across all classification networks are shown in the table's final column. All preceding averages were calculated without regard for the diagonal values, implying that they accurately show cross-dataset performance.

The penultimate row of the tables displays the standard deviation and average of the values in a block's diagonal, representing the performance of a single classifier on the same dataset (SD AVG). To show the classifier's performance across datasets, the final column of a table comprises the standard deviation and average of all non-diagonal results within a block (CD AVG).

In Table 4, we present experiments demonstrating that the HAN architecture is superior to the one [63] developed with the four best-performed machine learning approaches: discriminative (SVM and LR), decision trees (Random Forest), instance-based (KNN). The LS dataset achieves the highest AUC results across all the classifiers in the same dataset setup, indicating that it is straightforward to classify using Word2vec features. On the other hand, FT+HAN used in this study gives the highest result among all the classifiers. Training with dataset EN and testing with the LS dataset give the top result of **0.957** in the cross-dataset setup. On the other hand, training with dataset LS and testing with the GS dataset gives the worst result of **0.48** in the cross-dataset setup.

Table 5. Comparison of the proposed method with deep and machine learning studies

| Ref. | Method | Dataset | Accuracy | Precision | Recall | F1 | AUC |
|---|---|---|---|---|---|---|---|
| [45] | BERT+SVM | EN | - | 0.9772 | 0.9769 | 0.9770 | 0.9964 |
| | BERT+Logistic Regression | | - | 0.9786 | 0.9783 | 0.9784 | 0.9971 |
| | BERT+KNN | | - | 0.9654 | 0.9637 | 0.964 | 0.9905 |
| | BERT+Random forest | | - | 0.9639 | 0.9634 | 0.9635 | 0.9946 |
| [48] | BERT | SA | 0.98 | 0.96 | 0.99 | 0.9764 | - |
| | | EN | 0.97 | 0.96 | 0.98 | 0.9720 | - |
| | | LS | 0.98 | 0.90 | 0.98 | 0.9400 | - |
| [64] | SVM + LEP | EN | 93.9 | - | - | - | - |
| | | GS | 93.9 | - | - | - | - |
| This study | FT+HAN | TR | 0.992 | 0.989 | 0.999 | 0.994 | 0.999 |
| | | GS | 0.954 | 0.953 | 0.957 | 0.955 | 0.991 |
| | | SA | 0.955 | 0.893 | 0.978 | 0.933 | 0.987 |
| | | EN | 0.958 | 0.981 | 0.937 | 0.958 | 0.989 |
| | | LS | 0.980 | 0.933 | 0.948 | 0.940 | 0.997 |

According to SD AVRG results for the same dataset, our method outperforms all other methods with a value of 0.9926, which equates to a performance increase of at least 3.66% over [63]. Additionally, according to CD AVRG results corresponding to cross-dataset performance, our method outperforms all other methods with a value of 0.80645, which equates to a performance increase of at least 5.59% over [63].

In Table 5, we compare the results of the proposed study with recent deep learning and machine learning studies that utilize the same dataset. According to the results, the proposed method is comparable and, in some metrics, even outperforms the alternatives. This study surpasses the method the precision in [45] and the accuracy in [64]. In general, the study of [48] outperforms all other studies, but it is well known that the training or fine-tuning of the BERT model requires a great deal of time.

## 5. Conclusion

The FT+HAN architecture for email spam detection in a cross-dataset setting is proposed in this study. The approach combines CNN with attention techniques at both sentence and word levels. HAN advances spam detection accuracy by incorporating the email structure into the architecture and utilizing CNNs to extract more plentiful features. The proposed model architecture was evaluated using five distinct datasets, with cross-dataset and same-dataset comparisons used to determine the complexity of the latter. In the cross-dataset experiment, the training is performed on email data from one dataset and tested on data from another, taking into account that the datasets were gathered using distinct independent setups. This is done to simulate future unpredictable or variable conditions in email content distributions, such as those that might occur in a real-world scenario. When compared to previous works, our approach outperforms them in both same-dataset and cross-dataset evaluation.

**Statements and Declarations**

The authors declare that they have no conflict of interest.